\pdfoutput=1

\documentclass[11pt]{article}

\usepackage[final]{acl}

\usepackage{times}
\usepackage{latexsym}
\usepackage{multirow}
\usepackage{amssymb}
\usepackage{colortbl}
\usepackage{arydshln}
\usepackage{graphicx}
\usepackage{booktabs}
\usepackage{array}
\usepackage{soul}
\usepackage{tcolorbox}
\usepackage{wrapfig}
\usepackage{hyperref}
\hypersetup{
    colorlinks=true,
    linkcolor=blue,
    filecolor=magenta,      
    urlcolor=cyan,
    pdftitle={Overleaf Example},
    pdfpagemode=FullScreen,
    }
\usepackage[T1]{fontenc}

\usepackage[utf8]{inputenc}

\usepackage{microtype}

\usepackage{inconsolata}

\usepackage{graphicx}

\title{Precision or Recall? An Analysis of Image Captions for Training Text-to-Image Generation Model}

\author{
Sheng Cheng \quad \quad Maitreya Patel \quad \quad Yezhou Yang \\
Arizona State University \\
  \texttt{\{scheng53, maitreya.patel, yz.yang\}@asu.edu}}

\begin{document}
\maketitle
\begin{abstract}
Despite advancements in text-to-image models, generating images that precisely align with textual descriptions remains challenging due to misalignment in training data.
In this paper, we analyze the critical role of caption precision and recall in text-to-image model training. 
Our analysis of human-annotated captions shows that both precision and recall are important for text-image alignment, but precision has a more significant impact.
Leveraging these insights, we utilize Large Vision Language Models to generate synthetic captions for training. 
Models trained with these synthetic captions show similar behavior to those trained on human-annotated captions, underscores the potential for synthetic data in text-to-image training.\footnote{The data and code is available at \url{https://github.com/shengcheng/Captions4T2I}.}
\end{abstract}

\section{Introduction}
Recent advancements in diffusion models such as Stable Diffusion~\cite{Rombach_2022_CVPR}, DallE 3~\cite{betker2023improving}, Emu~\cite{dai2023emu}, and Imagen~\cite{saharia2022photorealistic}, have demonstrated remarkable capabilities in image synthesis.
Despite these achievements, challenges persist in generating images that accurately align with the given text inputs~\cite{huang2023t2i}.
One issue is the misalignment between training captions and images, where captions either describe only a portion of the image or fail to describe the image content accurately.
Recent research efforts have focused on enhancing caption quality using Large Language Models (LLM)~\cite{fan2024improving} or Large Vision Language Models (LVLM)~\cite{lai2023scarcity, chen2024pixartalpha}.
Nevertheless, there is a scarcity of in-depth analysis on how specific factors influence the efficacy of text-to-image model training.

\begin{figure}[!t]
    \centering
    \includegraphics[width=\columnwidth]{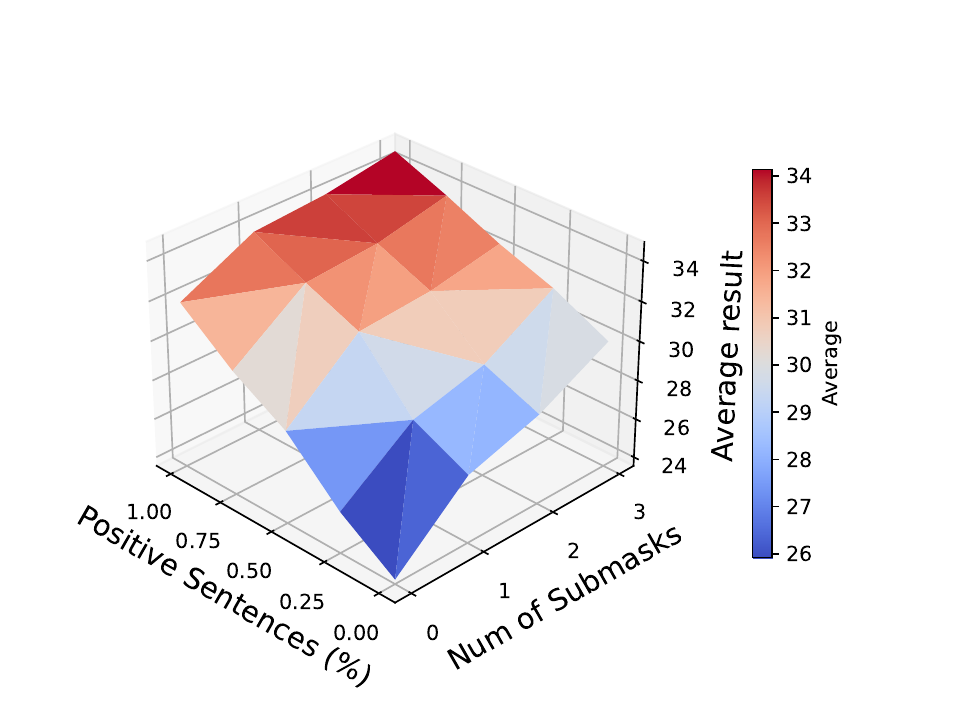}
    \caption{The result of the compositional capabilities across various combinations of precision and recall on human-annotated captions. Positive sentences indicate the precision of the captions, while the number of submasks represents the comprehensiveness of the captions.}
    \label{fig:score}
\end{figure}

In this paper, we evaluate captions based on two metrics: precision and recall. 
We train the text-to-image (T2I) model using human-annotated captions that vary in levels of precision and recall. 
We then assess the model’s capability for text-image alignment, specifically focusing on compositional ability.
As illustrated in Figure~\ref{fig:score}, our findings indicate that while combinations of high precision and high recall yield the best results, generating captions with high precision is generally more beneficial. However, if the model is prone to hallucination, adding more diverse details can also enhance performance. 

Building on these insights as shown in Fgiure~\ref{fig:score}, we explore whether the same observation hold when using LVLMs to generate synthetic captions for T2I training. Given the variability across different LVLMs, some models may emphasize diversity at the expense of precision, leading to increased hallucination, while others may prioritize precision but sacrifice diversity. We evaluate the precision and recall of captions generated by various LVLMs and then use them to train T2I models.
Our findings confirm that the compositional capabilities of the T2I models are consistent with our previous conclusions, underscoring the critical role of precision in caption generation.

The major contributions of this paper are:
\begin{itemize}
    \item We systematically evaluate the impact of precision and recall on T2I model training, establishing that while both metrics are important, precision has a more significant influence on the model’s performance.
    \item We extend our analysis by employing several LVLMs to generate synthetic captions. Our experiments show that the performance of T2I models trained with these synthetic captions is consistent with insights derived from human-annotated captions.
\end{itemize}

\section{Related Work}

\paragraph{Text-to-image diffusion model}
Given an input image $x$, it is paired with a corresponding caption $c$, which is segmented into multiple sentences $c_0, c_1, ..., c_N$. 
The sentence $c_0$ typically provides a general description of the image, while  $c_1, ..., c_N$ detail the characteristics of specific subregions within the image. 
Current state-of-the-art text-to-image generation model is the latent diffusion model~\cite{ho2020denoising, sohl2015deep, nichol2021improved, Rombach_2022_CVPR, peebles2023scalable}.
This model can be formulated as 
$$\mathcal{L} := \mathbb{E}_{\varepsilon(x),c,\epsilon\sim\mathcal{N}(0,1),t} \left[ \left\| \epsilon - \epsilon_\theta (z_t, t, \tau_\theta (c)) \right\|^2_2 \right],$$ 
where $\epsilon$ is the noise, $t$ denotes the denoising timestep, $\theta$ represents the parameters of the diffusion model, $\varepsilon$ and $\tau$ are image and text encoder. 

\paragraph{Improving captions for text-to-image model training}
Models such as Pixart-$\alpha$~\cite{chen2024pixartalpha}, DallE 3~\cite{betker2023improving}, and Stable Diffusion 3~\cite{esser2024scaling} emphasize the critical role of high-quality captions in their training. These systems incorporate captions synthesized by LVLMs~\cite{wang2023cogvlm, liu2024visual, 2023GPT4VisionSC} into their training processes to enhance T2I generation capabilities.
In particular, \cite{betker2023improving} investigates how synthetic captions contribute to improved T2I generation.
However, these models do not specifically examine how caption quality impact the effectiveness of their training processes.

\section{Analysis of Image captions for T2I training}

\paragraph{Dataset Construction} Our study utilizes the Dense Caption Dataset~\cite{urbanek2023picture}\footnote{CC-BY-NC license}, which comprises 8,012 images and 99,445 submasks derived from the SAM dataset~\cite{kirillov2023segment}.
Each image in the dataset is segmented into multiple submasks, and both the main image and its corresponding submask images, known as subimages, are each paired with at least one detailed, human-annotated caption. We refer to the captions associated with the main image as the main captions and those linked to the subimages as subcaptions.
These captions have been condensed to summarized versions of no more than 77 tokens using the LLAMA2-70B model~\cite{touvron2023llama}.
To create negative captions, the LLAMA2-70B model modifies these sentences by altering their structure, editing content, or reshuffling words to form new sentences with the original vocabulary.

Our dataset is constructed based on these elements.
In our study, we focus solely on how captions, including those for specific subregions within images, align with the overall image. 
Therefore, we do not use the subimages derived from submasks; instead, our dataset is built around the subcaptions associated with these submasks. 
To control the recall of the caption, each image is described not just by a basic caption that outlines the overall scene, but also by multiple subcaptions that provide detailed descriptions of specific regions, as shown in Table~\ref{tab:stat}.
We also manage the accuracy of our dataset by selectively including a certain proportion of negative captions or negative subcaptions.

\begin{table}[h]
    \centering
    \begin{tabular}{c c c c c}
    \toprule
    \# of submasks & 0 & 1 & 2 & 3 \\
    \midrule
    Avg. nouns & 12.3 & 20.5 & 28.0 & 35.2\\
    Avg. tokens & 54.8 & 91.0 & 125.1 & 157.8\\
    \bottomrule
    \end{tabular}
    \caption{Data constructed analysis. The more submasks indicates more comprehensive description. The number of nouns is computed by Spacy~\cite{spacy2}. The number of tokens is computed by tokinzer~\cite{raffel2020exploring}}
    \label{tab:stat}
\end{table}

\begin{table*}[ht!]
    \centering
    \begin{tabular}{cccccccc}
    \toprule
        \textbf{Positive} & \textbf{Num of} & \multicolumn{3}{c}{\textbf{Attribute Binding}} & \multicolumn{2}{c}{\textbf{Object Relationship}} & \textbf{Average} \\ 
        \textbf{Sentences}& \textbf{Submasks}& \textbf{Color $\uparrow$} & \textbf{Shape $\uparrow$} & \textbf{Texture $\uparrow$} & \textbf{Spatial $\uparrow$} & \textbf{Non-Spatial $\uparrow$} & \textbf{Result} \\ 
    \midrule
        0\% & 0 & 0.1842 & 0.3079 & 0.3125 & 0.0884 & 0.2954 & 0.2377 \\ 
        0\% & 1 & 0.2322 & 0.3382 & 0.3389 & 0.1380 & 0.3031 & 0.2701 \\ 
        0\% & 2 & 0.2610 & 0.3340 & 0.3620 & 0.1446 & 0.3052 & 0.2814 \\ 
        0\% & 3 & 0.2870 & 0.3496 & 0.3840 & 0.1727 & 0.3066 & 0.3000 \\ 
        \hdashline
        25\% & 0 & 0.2277 & 0.3042 & 0.3282 & 0.1230 & 0.2995 & 0.2565 \\ 
        25\% & 1 & 0.2801 & 0.3346 & 0.3620 & 0.1357 & 0.3047 & 0.2834 \\ 
        25\% & 2 & 0.2881 & 0.3356 & 0.3819 & 0.1529 & 0.3051 & 0.2927 \\ 
        25\% & 3 & 0.3203 & 0.3515 & 0.4193 & 0.1700 & 0.3078 & 0.3138 \\ 
        \hdashline
        50\% & 0 & 0.2855 & 0.3271 & 0.3740 & 0.1243 & 0.3020 & 0.2826 \\ 
        50\% & 1 & 0.3250 & 0.3583 & 0.4180 & 0.1611 & 0.3062 & 0.3137 \\ 
        50\% & 2 & 0.3244 & 0.3529 & 0.4256 & 0.1727 & 0.3078 & 0.3167 \\ 
        50\% & 3 & 0.3358 & 0.3607 & 0.4291 & 0.1850 & 0.3082 & 0.3237 \\ 
        \hdashline
        75\% & 0 & 0.3186 & 0.3402 & 0.3939 & 0.1350 & 0.3057 & 0.2987 \\ 
        75\% & 1 & 0.3387 & 0.3586 & 0.4443 & 0.1760 & 0.3078 & 0.3251 \\ 
        75\% & 2 & 0.3454 & 0.3679 & 0.4401 & 0.1784 & 0.3089 & 0.3281 \\ 
        75\% & 3 & 0.3581 & 0.3710 & 0.4503 & 0.1917 & 0.3095 & 0.3361 \\ 
        \hdashline
        100\% & 0 & 0.3437 & 0.3599 & 0.4351 & 0.1507 & 0.3086 & 0.3196 \\ 
        100\% & 1 & 0.3500 & 0.3865 & 0.4695 & 0.1745 & 0.3091 & 0.3379 \\ 
        100\% & 2 & 0.3567 & 0.3872 & 0.4676 & 0.1832 & 0.3094 & 0.3408 \\ 
        100\% & 3 & 0.3717 & 0.3892 & 0.4662 & 0.1986 & 0.3100 & 0.3471 \\ 
    \bottomrule
    \end{tabular}
    \caption{The result of the compositional capabilities across various combinations of precision and recall on human-annotated captions. It contains five categories: color, shape, texture, spatial, and non-spatial. Positive sentences indicate the precision of the captions, ranging from 0\% to 100\%, while the number of submasks represents the comprehensiveness of the captions, ranging from 0 to 3.}
    \label{tab:dci}
\end{table*}
\paragraph{Training and Evaluation} The family of Stable Diffusion models (v1.4, 1.5, and 2.1)~\cite{Rombach_2022_CVPR} integrates the CLIP~\cite{radford2021learning} text encoder, which is limited to processing 77 tokens. 
To overcome this limitation, we employed the Pixart-$\alpha$ model~\cite{chen2024pixartalpha}, which utilizes the T5~\cite{raffel2020exploring} text encoder capable of handling up to 512 tokens. 
We fine-tuned the model using LoRA~\cite{hu2022lora} over 10 epochs, with batch size of 32 and learning rate of 1e-4.
All experiments are run on an  A100 GPU.

To validate the model’s ability to generate images accurately aligned with the provided text, we use the T2I-Compbench~\cite{huang2023t2i}, which includes five tasks. The first three tasks evaluate the model’s capacity to accurately generate multiple objects in terms of correct color, shape, and texture. These tasks convert the text prompts into questions, which are tested using the BLIP2 model~\cite{li2023blip}. The spatial task examines the model’s understanding of spatial directives like ‘left’ and ‘right’ using an object detection algorithm~\cite{zhou2022simple}. The non-spatial task focuses on object interactions and employs the CLIP model~\cite{radford2021learning} to evaluate alignment.

\paragraph{Results}
Our results are shown in Table~\ref{tab:dci}. 
We quantify the precision of the captions by the percentage of positive sentences and assess their recall through the number of submasks. It is important to clarify that a negative sentence does not necessarily indicate complete irrelevance to the associated image. Typically, such a sentence is mostly accurate but includes a few incorrect elements.

\begin{table*}[!ht]
    \centering
    \begin{tabular}{cccccccc}
    \toprule
        \textbf{Caption} & \textbf{T2I} & \multicolumn{3}{c}{\textbf{Attribute Binding}} & \multicolumn{2}{c}{\textbf{Object Relationship}} & \textbf{Average} \\ 
        \textbf{Method}& \textbf{Model}&  \textbf{Color $\uparrow$} & \textbf{Shape $\uparrow$} & \textbf{Texture $\uparrow$} & \textbf{Spatial $\uparrow$} & \textbf{Non-Spatial $\uparrow$} & \textbf{Result} \\ 
    \midrule

    LLAVA & SD2.1 & 0.5406 &0.4310	&0.4941&	0.1370&	0.3165&	\textbf{0.3838}\\
    
    uform  &SD2.1 &0.5246	&0.3878&	0.5015&	0.1250&	0.3163&	0.3710\\
    BLIP &SD2.1 & 0.4962&	0.4189&	0.5087&	0.1457&	0.3156	&0.3770\\
    \hdashline
    LLAVA& SDXL & 0.4942 &	0.3942&	0.4782&	0.1452&	0.3143 & \textbf{0.3652}\\
    uform & SDXL &0.4496&	0.3458&	0.4130&	0.1185&	0.3116& 0.3277 \\
    BLIP & SDXL &0.4426	&0.3652&	0.4307&	0.1395&	0.3115& 0.3379\\
       
    \bottomrule
    \end{tabular}
    \caption{The result of the compositional capabilities on synthetic captions generated through different LVLMs.}
    \label{tab:sythn}
\end{table*}
Our experiments confirm that both precision and recall influence the compositional capabilities of the model. However, our findings indicate a more significant impact of precision on performance as compared to recall. 
Notably, models trained with 0\% positive sentences and three additional subcaptions underperform significantly relative to those trained with 100\% positive sentences, even in the absence of any subcaptions, which contain approximately four times less information.
At lower precision levels, increasing recall significantly boosts performance. For example, improving recall with captions that have 0\% precision results in a 6.3\% gain in performance. However, as the precision of captions improves, the benefits of increasing recall decrease. When captions are 100\% precise, the additional performance gain from increased recall is just 2.8\%.

\section{Insight for Synthetic captions for T2I training}

Building on the findings from our previous analysis, this section investigates whether the observed impacts of precision and recall on performance also apply to synthetic captions.

We conducted experiments using synthetic captions generated by three different LVLMs: LLAVA~\cite{liu2024visual}, BLIP2~\cite{li2023blip}, and uform\footnote{https://huggingface.co/unum-cloud/uform-gen2-qwen-500m}. 
Each model was given the instructions: \textit{“Describe the image concisely,”} with a limit of fewer than 77 tokens per caption. 
The images for training are sourced from the MSCOCO dataset~\cite{lin2014microsoft}, which includes 118k images. 
We fine-tuned the Stable Diffusion v2.1 and Stable Diffusion XL base 1.0~\cite{podell2023sdxl}\footnote{https://huggingface.co/stabilityai/stable-diffusion-xl-base-1.0} model using these captions for 100,000 iterations with a batch size of 8 and a learning rate of 1e-4.

Due to the high cost of human verification for the precision and recall of synthetic captions, we use a modified version of the Faithscore~\cite{jing2023faithscore} for evaluation. 
Initially, we filter out the descriptive content of the captions. 
Then, adapting from the original method in \cite{jing2023faithscore}, which decomposes captions into ENTITY, COUNT, COLOR, RELATION, and OTHER, we refine our decomposition to better fit the compositional evaluation metric by using ENTITY, SHAPE, COLOR, TEXTURE, SPATIAL, NON-SPATIAL, and OTHER. 
Each sentence within these categories is then assessed for correctness. For the first two steps, we utilize the GPT-3.5 API~\cite{brown2020language}, and for the final step of evaluation, we use the LLAVA1.5-13B model, following the approach detailed in \cite{jing2023faithscore}.
Due to the costs of the API usage, we randomly select 1000 samples to probe the quality of the captions. 
The result is presented in Table~\ref{tab:faithscore}, including the number of entities per caption, which represents the recall of the captions, and their corresponding faithscore, which indicates the precision of the captions.
It reveals that the BLIP model generates captions with less information but achieves high precision. Conversely, uform provides more diverse information but with relatively lower precision. Meanwhile, the LLAVA model not only maintains high precision but also exhibits better comprehensiveness compared to BLIP.

\begin{table}[!t]
    \centering
    \begin{tabular}{c c c c }
    \toprule
    Method & LLAVA & BLIP & uform \\
    \midrule
    \# of Entites & 4.90 & 2.12 & \textbf{6.52} \\
    Faithscore & 0.911 & \textbf{0.931}& 0.831 \\
    \bottomrule
    \end{tabular}
    \caption{The number of entities and  modified faithscore of synthetic captions generated from three different LVLMs model.}
    \label{tab:faithscore}
\end{table}

The results of the compositional capabilities of T2I models trained with synthetic captions are shown in Table~\ref{tab:sythn}. 
The findings reveal that the LLAVA model, which has relatively high precision and recall, outperforms the other two models. 
Despite containing three times less information than uform, the BLIP model’s high precision enables it to perform better than the uform model. 
This observation aligns with insights from human-annotated captions, affirming that high precision is more crucial than high recall.

\section{Conclusion}
In this study, we investigated how caption quality affects T2I model training. 
We found that while both precision and recall are important, precision is more crucial for effective training. 
These findings are confirmed using both human-annotated and synthetic captions from LVLMs. 
This insight could help improve the creation of synthetic captions for future T2I training.

\section*{Limitation}
A key limitation of our study is the use of the LLAVA1.5-13B model in the Faithscore evaluation to determine the correctness of each entity in the image. 
Since synthetic captions are also generated with the LLAVA model, our evaluation might inherently favor captions generated by it. 
However, the LLAVA model remains one of the most advanced open-source Vision-Language Models available. Additionally, the cost of using human annotation for evaluation would be significantly high. 
In future work, we plan to explore using GPT-4 for evaluation to reduce this bias potentially.

\section*{Acknowledgements}

This work was supported by NSF Robust Intelligence program grants \#1750082 and \#2132724. The authors acknowledge the resources provided by Research Computing at Arizona State University and the National Artificial Intelligence Research Resource (NAIRR pilot \#240117). The authors also acknowledge technical access and support from ASU Enterprise Technology.
The views and opinions of the authors expressed herein do not necessarily state or reflect those of the funding agencies and employers.

\bibliography{custom,anthology}

\appendix
\section{Visualization of Human-annotated captions}
We show sample that contain all positive sentences with 3 submasks (All-pos) and all negative sentences with 3 submasks (All-neg).
\begin{figure}[h!]
    \centering
    \includegraphics[width=\columnwidth]{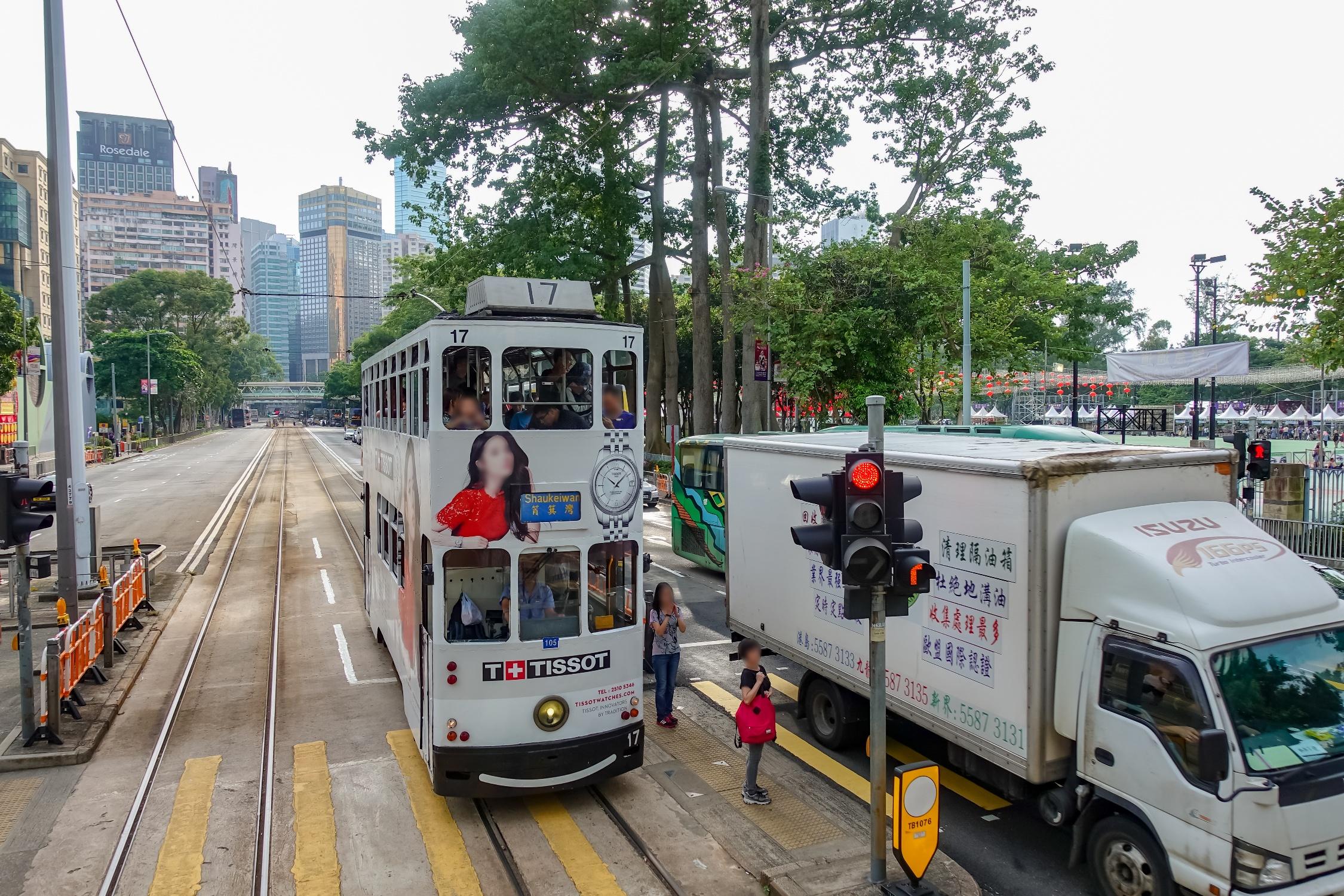}
    \caption{One sample from the DAC dataset, used for analysis of human-annotated captions.}
    \label{fig:hum1}
\end{figure}

All-pos: \textit{A white double-decker bus and truck are parked at an intersection, with a urban skyline in the background. The bus has a Tissot logo and watch, while the truck has Chinese writing and phone numbers. The intersection has traffic lights, pedestrian crossing button, and yellow tactile paving for visually impaired people. A white Isuzu truck is parked on the right side, with a red 166PS logo on the roof and a driver visible through the right side door. There are four lines of Chinese words in red, blue, and green on the trunk, and three phone numbers at the bottom. Overcast sky with a darker gloom on the left. A white double-decker bus with a Tissot logo and watch picture under the windshield, an actress in red clothing, and a driver in blue uniform. The bus has 17 written on top and bottom, and people can be seen sitting and standing inside through the windows.}

All-neg: \textit{A sleek black double-decker bus and a rusty old truck are parked at a bustling roundabout, with a picturesque countryside landscape in the background. The bus features a large advertisement for a luxury fashion brand, while the truck is covered in colorful graffiti. The roundabout has a central fountain, lined with benches and surrounded by vibrant flower beds. A white Isuzu truck with a red logo on the roof has a driver visible through the right side door. The trunk features a colorful design with four lines of words, while three phone numbers are displayed at the bottom. The engine rumbles, indicating a powerful 166PS output. A blue sky with a hint of clouds. A white double-decker bus with a Tissot logo, an actress in yellow, and a driver in navy. The bus has 19 written on it and people are partying inside with streamers and balloons.}

\section{Visualization of Synthetic captions}
We show sample that contain the synthetic captions from LLAVA, BLIP, and uform.

\begin{figure}[ht]
    \centering
    \includegraphics[width=\columnwidth]{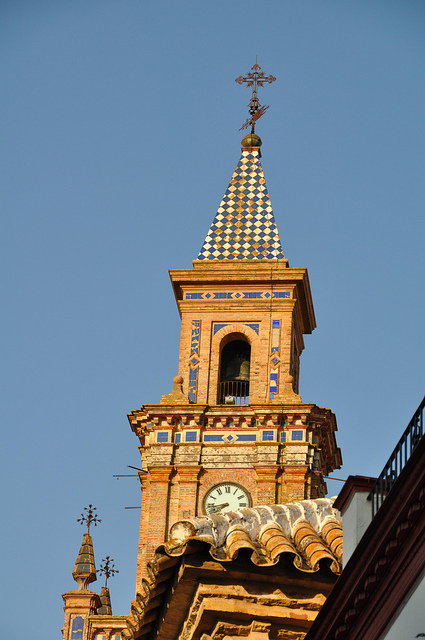}
    \caption{One sample from MSCOCO, used for generating synthetic captions by LVLM.}
    \label{fig:syth1}
\end{figure}

LLAVA: \textit{The image features a tall clock tower with a blue and gold design. The tower is adorned with a cross on top, adding a religious touch to the structure. The clock is positioned towards the center of the tower, making it a prominent feature. The tower stands out against a blue sky.}

BLIP: \textit{a clock tower on a building with a clock on top.}

uform: \textit{A majestic brick clock tower with a blue and white tiled roof stands tall against a clear blue sky, featuring a cross at the top and a bell at the bottom. The tower is surrounded by other buildings, creating a serene urban landscape.}

\end{document}